\documentclass[conference]{IEEEtran}
\IEEEoverridecommandlockouts
\usepackage{cite}
\usepackage{amsmath,amssymb,amsfonts}
\usepackage{algorithmic}
\usepackage{graphicx}
\usepackage{textcomp}
\usepackage{xcolor}
\usepackage{algorithm2e}
\usepackage{mathtools}
\usepackage{subcaption}
\usepackage{tikz}
\DeclareMathOperator{\sign}{sgn}

\def\BibTeX{{\rm B\kern-.05em{\sc i\kern-.025em b}\kern-.08em
     T\kern-.1667em\lower.7ex\hbox{E}\kern-.125emX}}
\begin{document}


\title{Confident in the Crowd: Bayesian Inference to Improve Data Labelling in Crowdsourcing\thanks{This work is based on the research supported in part by the National Research Foundation of South Africa (Grant Number: 118075).}}

\author{\IEEEauthorblockN{Pierce Burke\IEEEauthorrefmark{1} and
Richard Klein\IEEEauthorrefmark{2}}
\IEEEauthorblockA{School of Computer Science and Applied Mathematics,\\
University of the Witwatersrand, Johannesburg, South Africa\\
Email: \IEEEauthorrefmark{1}pburke1172@gmail.com,
\IEEEauthorrefmark{2}richard.klein@wits.ac.za}}



\maketitle


\begin{abstract}
With the increased interest in machine learning and big data problems, the need for large amounts of labelled data has also grown. However, it is often infeasible to get experts to label all of this data, which leads many practitioners to crowdsourcing solutions. In this paper, we present new techniques to improve the quality of the labels while attempting to reduce the cost. The naive approach to assigning labels is to adopt a majority vote method, however, in the context of data labelling, this is not always ideal as data labellers are not equally reliable. One might, instead, give higher priority to certain labellers through some kind of weighted vote based on past performance. This paper investigates the use of more sophisticated methods, such as Bayesian inference, to measure the performance of the labellers as well as the confidence of each label. The methods we propose follow an iterative improvement algorithm which attempts to use the least amount of workers necessary to achieve the desired confidence in the inferred label. This paper explores simulated binary classification problems with simulated workers and questions to test the proposed methods. Our methods outperform the standard voting methods in both cost and accuracy while maintaining higher reliability when there is disagreement within the crowd.


\end{abstract}


\section{Introduction}
Large amounts of data have become an integral part of many different fields in science, however, the data on its own is not always meaningful and often need extra labels along with it. Due to the scale of the data, it is often infeasible for it all to be labelled by one person or even a small team. For this reason, many machine learning practitioners are turning to crowdsourcing platforms for low-cost data labelling. Crowdsourcing has been shown to solve problems across many different fields \cite{brabham2008crowdsourcing}. A real-world example of when crowdsourcing was useful was shown in a paper by Foody \textit{et al}. \cite{foody2013assessing} where they used volunteer labels on geographical images to determine land cover in certain areas. ImageNET is an image dataset which consists of over 14 million labelled images which are used for training computer vision models. A large portion of ImageNET was outsourced to online workers through an online platform called Amazon Mechanical Turk \cite{deng2009imagenet}.

To ensure that the quality of the labels is not biased by a single workers aptitude, it is beneficial to ask multiple workers to provide labels for the data \cite{sheng2008get}. This introduces the question of what is the best way of aggregating these workers labels to determine the true label. One of the most common methods for determining the ground truth label from a set of labels is by taking the majority vote \cite{sheng2008get,snow2008cheap}. While this method is often used across many scenarios, it opens up the issue of having to use many workers to ensure the quality of the label is not compromised by unreliable workers.

Another method which tries to address the issue of unreliable workers is by weighting workers based on some reliability metric. A common way of determining the reliability of a new worker is by giving a test question for which we already know the ground truth and compare their answer to what we know to be true \cite{le2010ensuring}. This method fails to account for the possible change in workers reliability or the chance that their ``test'' questions were uncharacteristic of them. Instead, a better method would be one which can continuously learn and update new information about the workers. 
In this paper, three possible methods will be explored which try to solve these problems. The first one will be a weighted majority vote scheme that assigns the workers historical accuracy as the weight for their answer. The other two will be probabilistic methods that estimate the probability of a worker giving the correct label given their historical accuracy. Both a maximum likelihood and Bayesian inference model will be explored. The worker’s historical accuracy is judged by comparing their answers to the crowd consensus. 

The different methods are judged based on how accurately they predict the correct label for a binary classification task. 

This paper is structured as follows: Section 2 consists of related work and some of the other methods which have been proposed in this field; section 3 presents the formulation of the methods and design choices we have used and section 4 covers the results. Section 5 concludes our findings and presents future work.

\section{Background and Related Work}
Much work has been done to improve the quality of labels for a given task. In a method proposed by Jung and Lease \cite{jung2011improving}, a z-score metric to filter out unreliable workers is used. In another paper proposed by Kumar and Lease \cite{kumar2011modeling} they looked at comparing a single labeller method, a multi-labeller method with a majority vote and a multi-labeller method with Na\"{i}ve Bayes to improve label accuracy. However, the paper assumes that the worker accuracies were known to the system. Tarasov \textit{et al}. \cite{tarasov2014dynamic} looked at an approach for the dynamic estimation of rater reliability, specifically for regression tasks and they theorise that it could be applied to different types of problems such as multi-class classification. This method suggests that the problem can be formulated as a multi-armed bandit problem which needs to choose between exploiting current known workers skills or exploring new workers.
There have also been a few papers which model the workers or labels as a probabilistic distribution with latent variables. They then optimize the parameters of the distribution using Expectation Maximization \cite{raykar2010learning,whitehill2009whose,tang2011semi}. Raykar \textit{et al}. \cite{raykar2010learning} use a Bayesian model by imposing a prior on the worker's ability. However, they don't use full Bayesian inference but rather take a point estimate from the mode of the posterior.
In a paper put forward by Foody \textit{et al}. \cite{foody2013assessing} they used a latent class analysis method for estimating the quality of the labels given by the workers. 

\section{Research Methodology}
\subsection{Data}
\subsubsection{Workers}
All of the workers and questions asked in this paper were simulated. The system was set up such that each worker has a unique probability of getting a given question correct, each of these probabilities was drawn from a normal distribution, any samples which are smaller than zero or larger than one are rounded to zero and one to ensure it is a valid probability. The workers were assumed to belong to one of three classes: adversarial, normal and expert. The adversaries probabilities were drawn from a normal distribution with mean of 15 and a standard deviation of 5. The normal worker's probabilities were drawn from a distribution with mean 60 and a standard deviation of 15, this pool is assumed to be the largest while also have the biggest variances between workers abilities. This leads to a situation where some workers can be ``hidden'' experts in the pool. The expert's probabilities are drawn from a normal distribution with mean 85 and standard deviation 5. These latent probabilities were not known to the different models. The workers in the expert pool are known to the system but come at a higher cost. The cost of asking an expert is five times the cost of asking a worker and the worker was taken as unit cost -- although this hyperparameter could be adjusted for different situations.
\subsubsection{Questions}
To simulate the varying difficulty in questions, a ``difficulty’’ value is added for each question. A higher difficulty rating will decrease the probability of a normal/expert worker getting the answer correct and decrease the probability of an adversarial worker getting the answer incorrect. This is under the assumption that adversarial workers are trying to give misleading results by answering what they think is incorrect. The probability is capped at 50\% for hard questions, as this represents the worker completely guessing. The set of difficulty values used to simulate the set of questions were drawn from a normal distribution with varying means and a standard deviation of 15.

\subsubsection{Available workers}
To emulate the varying availability of people, the entire worker pool is not available at all times. Instead, the system will receive a randomly generated subset of available workers for each question, and they can query any of those workers for classification on a label. If the entire set of workers is available for each question then it would be beneficial to include a method which will occasionally use non-optimal workers to ensure we explore all workers abilities. 

\subsection{Majority vote}
The first method looked at was the common majority vote. For each question, a set of available workers were asked to give a label and then the responses were combined by choosing the label that appeared the most. To avoid the possibility of a tie, only an odd number of workers were asked.

\subsection{Weighted majority vote}
The weighted majority vote method tries to improve accuracy by weighting the better workers higher. The weighting of each worker is calculated from the worker’s previous agreement with consensus. Calculation of label L from a set of predictions and workers weightings can be given by:
\begin{equation}
L = \sign\left({\sum_i v_i*w_i}\right).
 \label{Weighted}
\end{equation}
Where $v_i$ $\in[0,1]$ denotes the $i$'th workers weighting, $w_i$ $\in \{-1;1\}$ denotes the $i$'th workers prediction and sgn is the sign function.
\eqref{Weighted} will return a label L of either -1 or 1.
We calculate $v_i$ by estimating the worker's accuracies based on how much they have agreed with the consensus. For a given worker $i$, the respective weighting can be calculated as:
\begin{equation}
v_i = \frac{c_i}{N_i},
\end{equation} 
where $c_i$ denotes the number of questions worker $i$ has agreed with the previous consensus and $N_i$ is the total number of questions worker $i$ has answered. 

\subsection{Expectation Maximization}
In the previous method the expected accuracy of our workers is used to weight them. However, this can be expanded to a more probabilistic approach by considering the likelihood of a worker giving a label $w_i \in \{0,1\}$ given that the true label is $L = 1$. To model the likelihoods of the workers we will use the Bernoulli distribution. However, the latent variable of the worker's probability of giving the correct classification is unknown at the beginning. 
Instead, the parameters of maximum likelihood can be approximated at each iteration using the Expectation-Maximization (EM) algorithm.  The EM algorithm will consist of two steps, the expectation step where the label is calculated using the currently estimated parameters and then the maximization step which updates the parameters based on the previous data and the new data point added by the expectation step. These two steps are then repeated for each new question asked allowing for the system to learn more about the workers as more questions have been answered by each worker. This version of the EM differs from the methods used in \cite{raykar2010learning,whitehill2009whose,tang2011semi} as our implementation uses the EM to iteratively learn about the workers after giving them new questions and previous work optimises over an already available set of labels for a set of tasks. \newline
We make two assumptions:
\begin{enumerate}
    \item The workers are independent
    \item The prior probability of a label belonging to each class is uniform, i.e. $P(L=1) = P(L=0)$.
\end{enumerate}
Let $\lambda_i$ be the probability that worker $i$ will provide the correct label. Therefore,
\begin{equation}
P(w_i = L) = \lambda_i^{1-|w_i - L|} (1-\lambda_i)^{|w_i - L|},
\end{equation} and
\begin{eqnarray}
P(w_i | L = 1) & = & \lambda_i^{w_i} (1-\lambda_i)^{1-w_i},\label{eq:lam1}\\
P(w_i | L = 0) & = & \lambda_i^{1-w_i} (1-\lambda_i)^{w_i},\label{eq:lam2}
\end{eqnarray} where $w_i$ is worker $i$'s response and $L$ is the correct label. This assumes that the worker is equally likely to supply the correct label regardless of the true label. This assumption could be easily relaxed by using separate Bernoulli Distributions in \eqref{eq:lam1} and \eqref{eq:lam2}. 

Over multiple independent questions, the maximum likelihood estimate for $\lambda_i$ is: \begin{equation}
\lambda_i = \frac{c_i}{N_i},
\end{equation} where $c_i$ denotes the number of times worker $i$ has agreed with the  consensus and $N_i$ is the total number of questions worker $i$ has answered \cite{princeCVMLI2012}.

The problem exhibits symmetry around the true label value. So without the loss of generality we consider the problem where the true label is $1$.

Bayes rule yields:
\begin{equation}
P(L = 1| w_{i...I}) = \frac{P(w_{i...I}| L = 1) P(L=1)}{\sum_{k=0}^1 P(w_{i...I}| L = k) P(L=k)}.
\end{equation}

Assumption 1 allows for the joint probability between workers to be changed into the product of probabilities.
\begin{equation}
P(L=1|w_{i...I}) = \frac{\prod_{i=0}^I P(w_i | L=1) P(L=1)}{\sum_{k=0}^1 \prod_{i=0}^I P(w_i | L=k) P(L=k) }
\end{equation}
Assumption 2 means that the prior probabilities in the numerator and denominator cancel each other out. This gives the new formula of
\begin{equation}
\small
P(L=1|w_{i...I}) = \frac{\prod_{i=0}^I P(w_i | L=1)}{ \prod_{i=0}^I P(w_i | L=1) + \prod_{i=0}^I P(w_i | L=0)},
\label{eq:likelihood}
\end{equation}
where $w_i$ is the label that worker $i$ provided.

Equation \eqref{eq:likelihood} gives us a probabilistic formula for calculating the likelihood of a label. The posterior allows us to measure the confidence associated with each label and is calculated given our current set of worker supplied labels. This confidence can then be used to determine how many workers are required for each question by adding more workers only if the confidence is below some threshold. The process to answering a single question is shown in Algorithm \ref{algo:iterative}.

The confidence of a label being correct given our workers responses, denoted by $\omega_{n}$, can be calculated by:
\begin{equation}
    \omega_{n} = |P(L=1|w_{i...I}) - (1 - L)|,
\label{confidenceCalculate}
\end{equation}
where $w_{i...I}$ is the set of all labels returned from each worker and $L$ is the label chosen by the system based on the probability calculated. The above formula will calculate how close the probability was to the predicted label for each respective label.

\begin{algorithm}
\SetAlgoLined
\KwResult{Label L}
 Get each workers prediction from initial set of workers\;
 Calculate the probability of label being 0 or 1 using Bayes formula and workers' parameters\;
 Calculate the confidence\;
 \While{Confidence $\leq$ threshold}{
  Get another worker's prediction\;
  Recalculate the probability\;
  Recalculate confidence\;
 }
 \If{Confidence $\leq$ threshold}{
   Calculate label using experts\;
   }
   Update workers' parameters based on the label\;
   return label\;
 
 \caption{Iterative label improvement}
 \label{algo:iterative}
\end{algorithm}

This provides a way to improve the answer iteratively by adding more workers as needed, instead of using all of the available workers.

\subsection{Bayesian Inference}
In the maximum likelihood approach discussed above, a point estimate for $\lambda_i$ is used to model each worker. This can cause over-confidence in the measure of a worker's skill based on previously seen data. As a way of reducing the effect of this over-fitting on the initial data a more conservative method can be used, such as Bayesian Inference. In this case, a prior over our workers' possible $\lambda$'s is introduced. To simplify the calculation of the probability density, we use a Beta distribution, which is conjugate the Bernoulli distribution\cite{princeCVMLI2012}. This can be given by:
\begin{equation}
P(w_i | L=1) = \int_{0}^1 \mathrm{Bern}_{w_i}[\lambda_i] \mathrm{Beta}_{\lambda_i}[\alpha_i , \beta_i] d\lambda_i .
\label{bayinf1}
\end{equation} 

Due to the conjugacy between the Bernoulli and Beta distributions \eqref{bayinf1} can be simplified to\cite{princeCVMLI2012}:
\begin{equation}
P(w_i | L = 1) = \frac{\Gamma[\alpha_i + \beta_i] \Gamma[\alpha_i + w_i] \Gamma[1 - w_i + \beta_i]}{\Gamma[\alpha_i]\Gamma[\beta_i] \Gamma[\alpha_i + \beta_i +1]}.
\label{bayDensity}
\end{equation}

Substituting \eqref{bayDensity} into \eqref{eq:likelihood} gives us the likelihood of the true label which can be used to calculate the predicted label in the same way as the previous method.

In \eqref{bayDensity}, the $\alpha$ and $\beta$ can be interpreted as prior parameters that describe a probability distribution over $\lambda_i$, the probability of the worker assigning the label correctly. This allows the system to track how confident it is in the estimate of each worker's skill. The larger $\alpha$ is relative to $\beta$, the larger the expected $\lambda_i$ will be from the Beta distribution.

To allow for the system to learn more about the workers as they answer questions, the $\alpha$ and $\beta$ parameters should update as more questions are answered by each of the workers. It can be shown from the above equations that the $\alpha$ and $\beta$ for the $n$th question are $\alpha + \sum_{i=1}^{n-1} c_i$ and $\beta + \sum_{i=1}^{n-1} (1-c_i)$,
with $c_i = 1$ if the worker agreed with consensus for question $i$ and $0$ otherwise.
This can be turned into an update rule by storing the $\alpha$ and $\beta$ for each worker and updating it after each question they answer.
\begin{eqnarray}
\alpha_{n+1} & = & \alpha_{n} + c_{n} \label{eq:a}\\
\beta_{n+1}  & = & \beta_{n} + 1 - c_{n}.\label{eq:b}
\end{eqnarray}\newline
Again, we assume symmetry in the problem -- that the workers are equally skilled regardless of whether the true label is $0$ or $1$. This assumption is made because we are trying to model the general case where no class is more difficult to label than another. When this is not the case, separate prior parameters for each case should be modelled.

This method follows the same Expectation Maximization approach as the maximum likelihood case except the prior parameters of the Beta Distribution are updated instead of the point estimate of $\lambda_i$.

\subsection{Bayesian inference with confidence update}
An adjustment to the method above can be made by changing the update rule of the $\alpha$ and $\beta$ parameters in \eqref{eq:a} and \eqref{eq:b}. Instead of updating by a binary value of $1$ or $0$, the update could incorporate information about how confident the system is in that label. This allows for small adjustments on labels which had a lot of disagreement or uncertainty and large adjustments based on labels in which there is high confidence. The update rule then becomes:
\begin{eqnarray}
\alpha_{n+1} = \alpha_{n} + \omega_{n} c_{n} \\
\beta_{n+1} = \beta_{n} + \omega_{n} (1 - c_{n})
\end{eqnarray}

Where $\omega_{n}$ is given in \eqref{confidenceCalculate}.



\section{Experiments}\label{sec:experiments}
In this section, we will discuss some of the results we have obtained from our tests. From here on the methods will be denoted as MV for majority vote, Weighted for the weighted majority vote, EM for the maximum likelihood method, BAY for the standard Bayesian inference method and CONF for the Bayesian inference method with the confidence update rule. The cost measured in the experiments relates to how many workers were used for each question. Using one worker has a cost of one and using one known expert has a cost of five. The shaded areas in the graphs represent the variance in results between the different experiments.

\subsection{Number of questions}
In this experiment, we tested how the different methods evolved as the workers were asked more questions. The confidence threshold was fixed at 0.9, the number of workers at 35 and the worker priors at 0.6 (for the Bayesian methods we set the beta distributions means to 0.6). This means the system expects the workers to have an average accuracy of 0.6 for questions with a difficulty value of 0. 


\begin{figure}[!b]
\centering
\begin{subfigure}[b]{0.365\textwidth}
    \includegraphics[width=0.885\columnwidth]{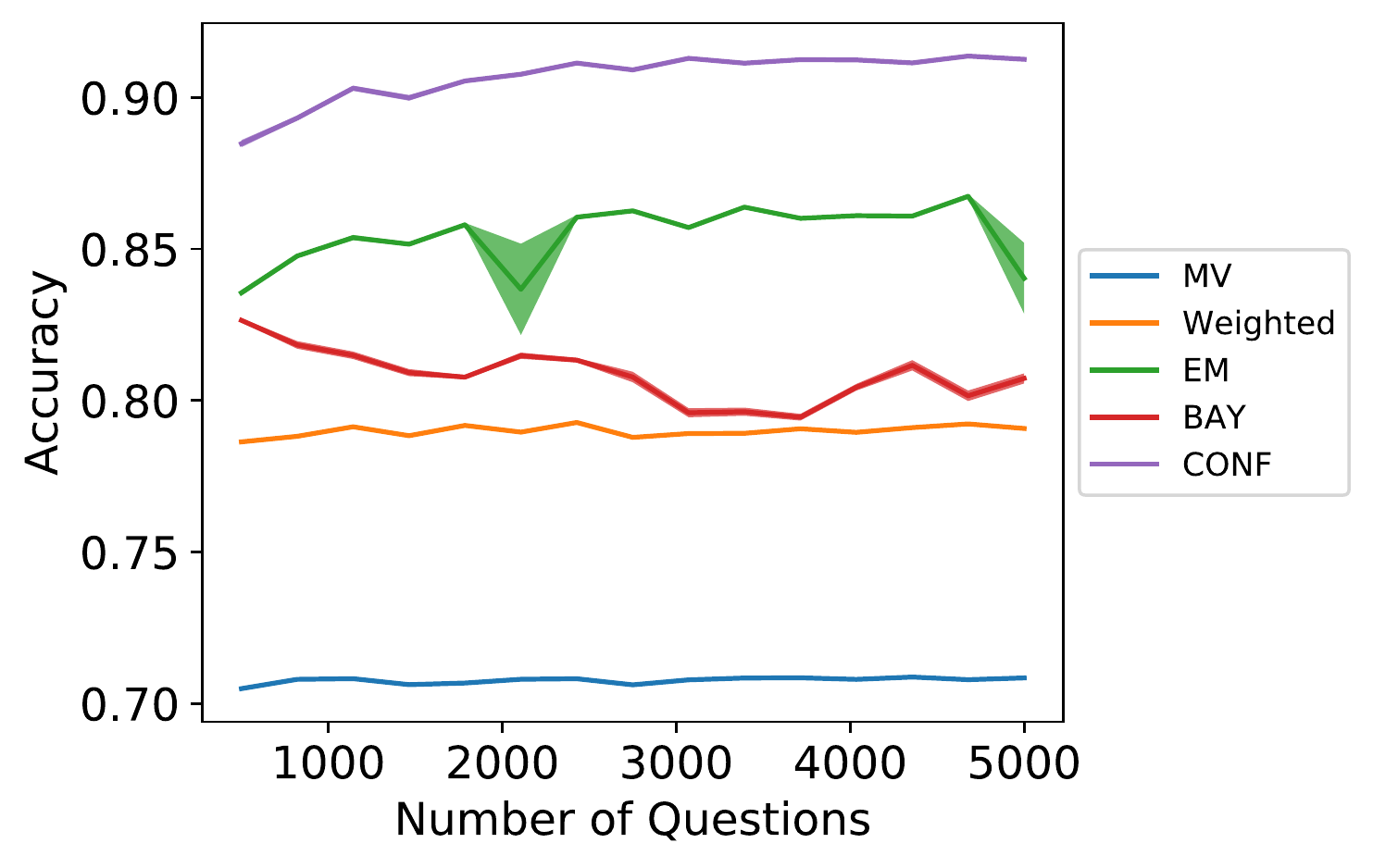}
    \label{fig:Questionsa}
    \caption{Accuracy}
\end{subfigure}
\begin{subfigure}[b]{0.365\textwidth}
     \includegraphics[width=0.885\columnwidth]{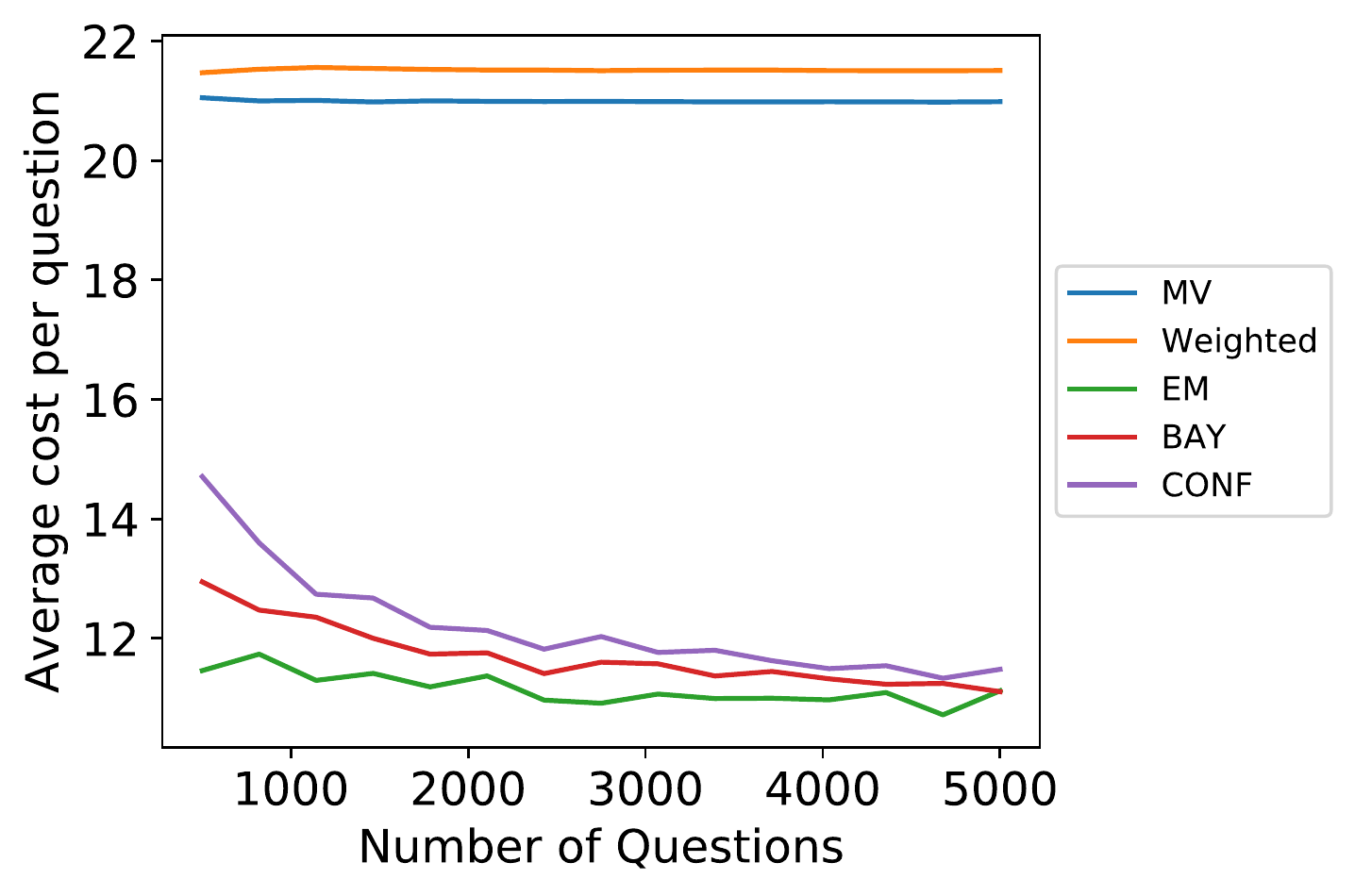}
    \label{fig:Questionsb}
    \caption{Average Cost}
\end{subfigure}
\caption{Number of Questions Test}
\label{fig:Questions}
\end{figure}

From Fig. \ref{fig:Questions}, the proposed methods outperform the majority vote method in both cost and accuracy. The majority vote implementation averages 71\% accuracy with an average cost of 21 per question. The weighted majority vote increases the accuracy to 78\% for approximately the same cost. The reason there is a discrepancy between the costs in these two methods is the majority vote will only use an odd number of workers, whereas the weighted majority vote method will use all of them. The probabilistic methods give even more improvement over the weighted majority vote with the worst one performing 3\% better and the best with an accuracy increase of 10\% on average. However, a major benefit that the probabilistic methods provide over the normal voting methods is the improved average cost. From the same figure, it can be seen that the cost of the probabilistic methods has a decreasing trend as the workers are asked more questions, this can be attributed to the fact that these models learn more about the workers as more questions get asked and they have a way of judging how confident they are in their answers. Out of the three probabilistic methods, the two Bayesian inference models have the highest average cost compared to the maximum likelihood, which shows how they are more conservative in their belief about the workers with the confidence update rule method being the most conservative. This conservativeness leads them to ask more workers per question to meet the confidence threshold. Because of the less conservative nature of the maximum likelihood method, there is greater variance in its results. This can be seen by the larger shaded areas.   

\subsection{Confidence Threshold}
For this test, the effect of different confidence thresholds on the probabilistic methods is explored. Each model was tasked with answering 2000 questions and the rest of the parameters were the same as in the previous test. 


\begin{figure}[!b]
\centering
\begin{subfigure}[b]{0.365\textwidth}
    \includegraphics[width=0.885\columnwidth]{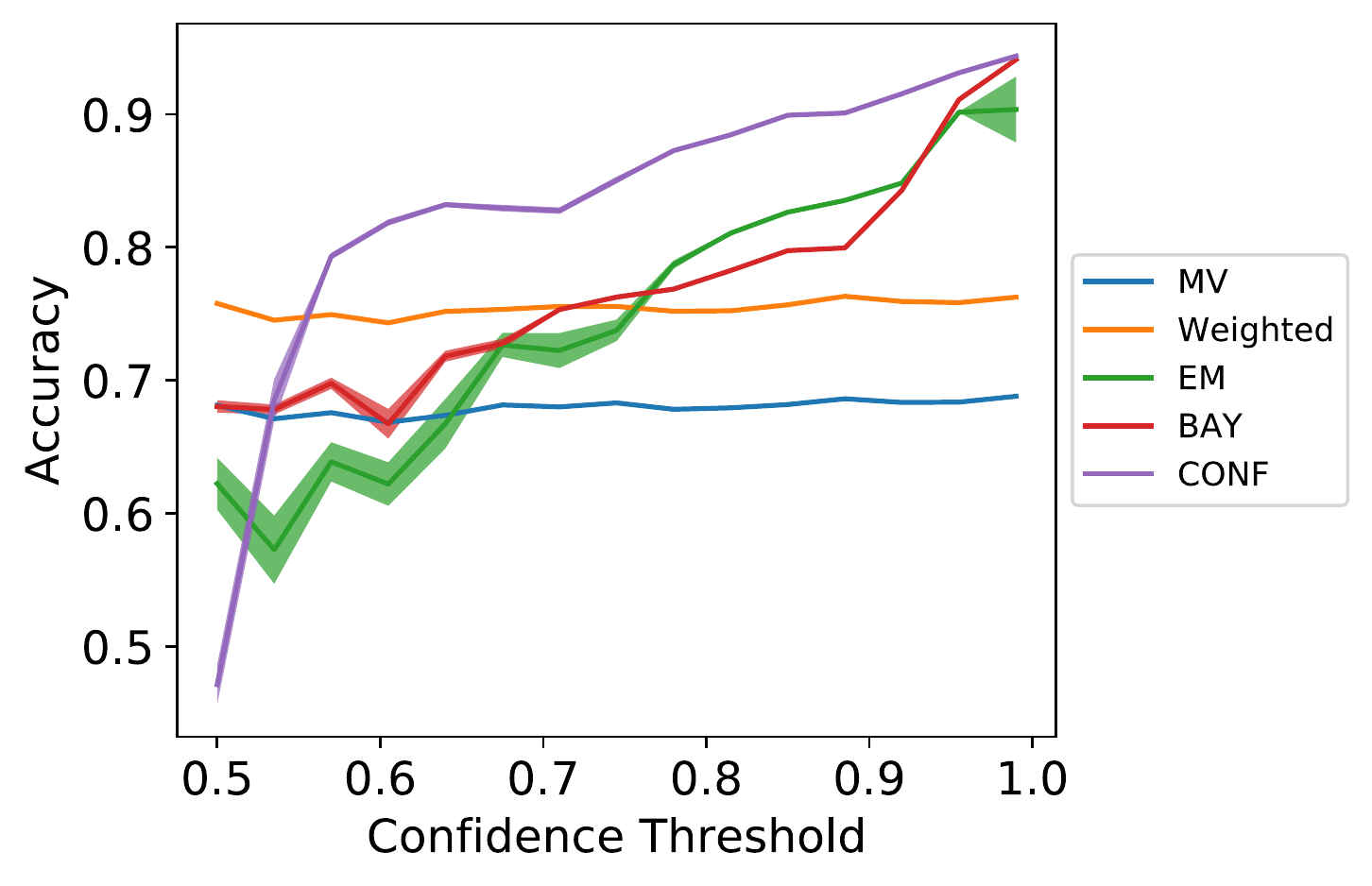}
    \label{fig:Confidencea}
    \caption{Accuracy}
\end{subfigure}
\begin{subfigure}[b]{0.365\textwidth}
     \includegraphics[width=0.885\columnwidth]{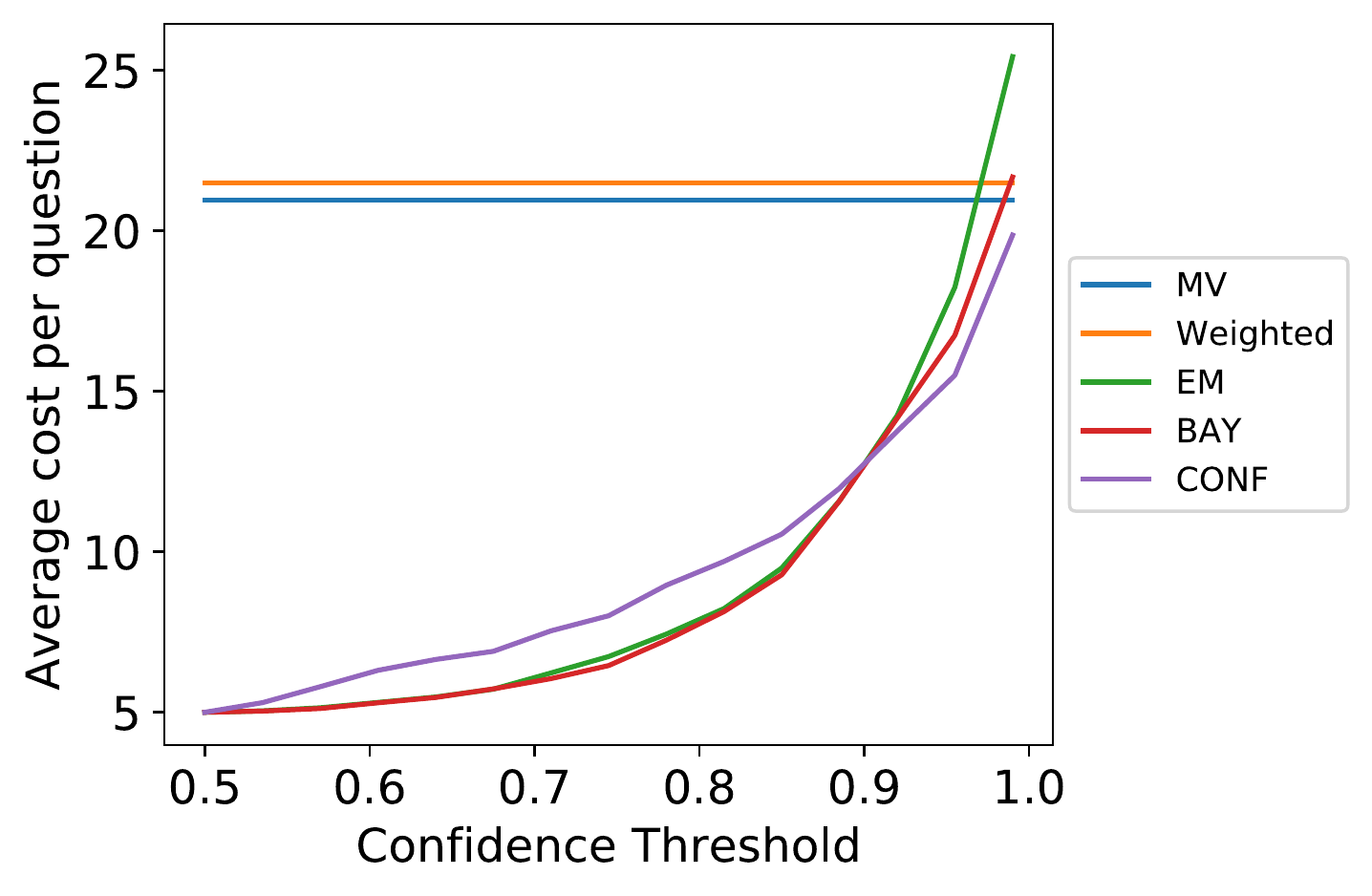}
    \label{fig:Confidenceb}
    \caption{Average Cost}
\end{subfigure}
\caption{Confidence Threshold Test}
\label{fig:Confidence}
\end{figure}

In Fig. \ref{fig:Confidence} it can be seen that when the confidence threshold is set at 0.5 the probabilistic methods perform even worse than the majority vote and weighted majority vote. A confidence value of 0.5 implies that each label is equally likely given our worker labels. This leads the probabilistic methods to take whatever label the initial set of workers gives them. The Bayesian inference with confidence update shows the lowest results for the low confidence threshold as the $\alpha$ and $\beta$ parameters of the beta distribution will change very little relative to each other. An interesting result that can be taken from the graph is that as the confidence threshold gets closer to 1, the two Bayesian inference models’ accuracies tend towards each other. This is most likely because the update rules for parameters of the two methods become more similar as the average confidence increases. If we look at the extreme case where the confidence was always 1, then the update rules would be equivalent. Increasing the confidence threshold had a drastic impact on the cost required, with the maximum likelihood method requiring higher cost than both the majority vote and weighted majority vote. From these results, we can conclude that for tasks which require lower confidence, it is more beneficial to use the standard Bayesian inference update rule or the maximum likelihood model and, when higher confidence is required, switching over to the confidence update rule is more beneficial.

\subsection{Difficulty}
This experiment compared how the question difficulty influenced the accuracy and costs of each method.
In our system, we simulate the varying difficulty of a question by increasing or decreasing the probability of a worker returning the correct label. When the difficulty is negative, the probability of returning the correct answer will increase and if it is positive it will decrease (If the worker is adversarial it will be the opposite). The probability changes will not go below 50\% for standard workers or above 50\% for adversarial. The difficulty of this test was varied between -15 and 15 to represent easier and more difficult questions respectively.


\begin{figure}[!b]
\centering
\begin{subfigure}[b]{0.365\textwidth}
    \includegraphics[width=0.885\columnwidth]{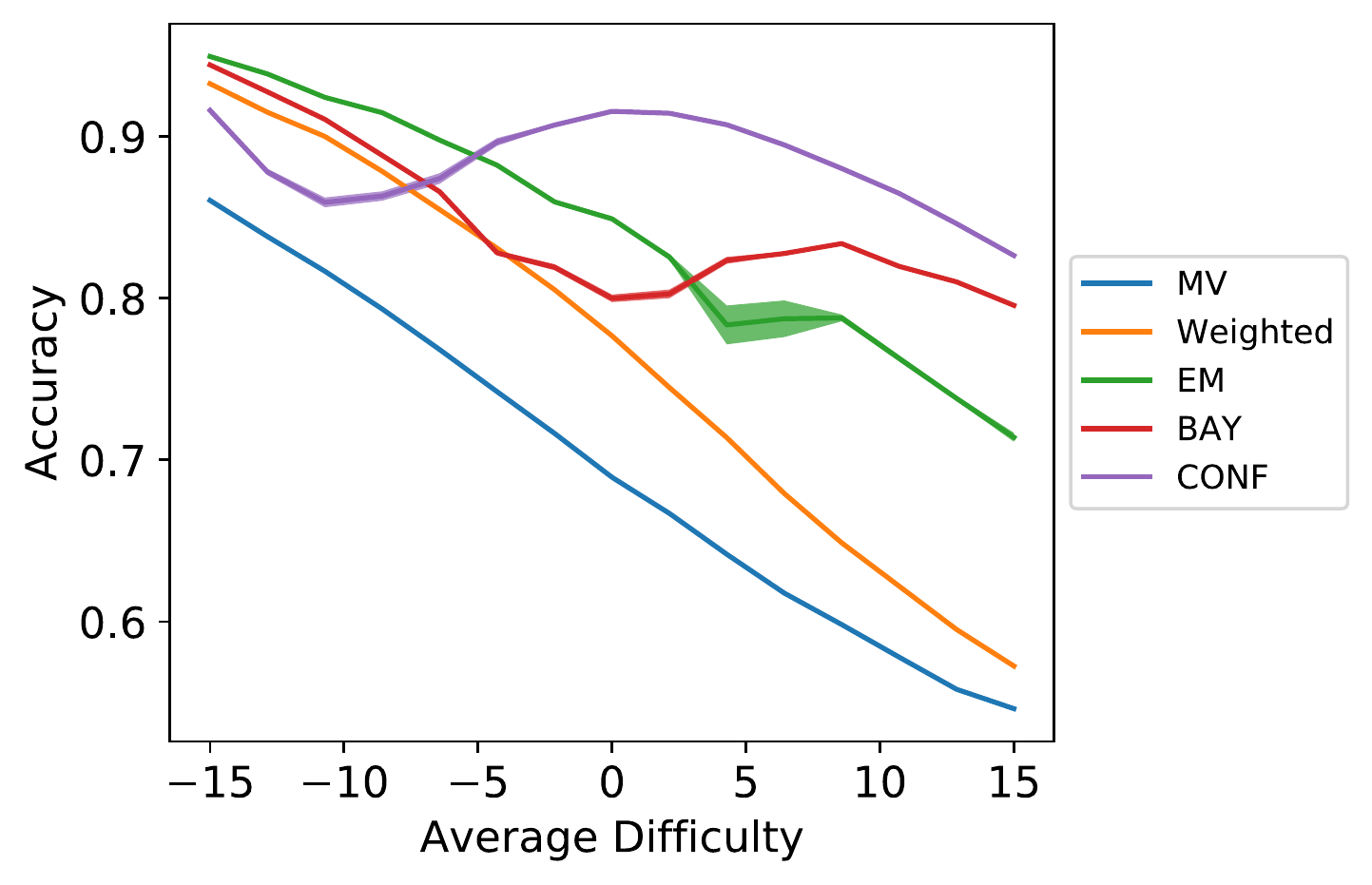}
    \label{fig:Difficultya}
    \caption{Accuracy}
\end{subfigure}
\begin{subfigure}[b]{0.365\textwidth}
     \includegraphics[width=0.885\columnwidth]{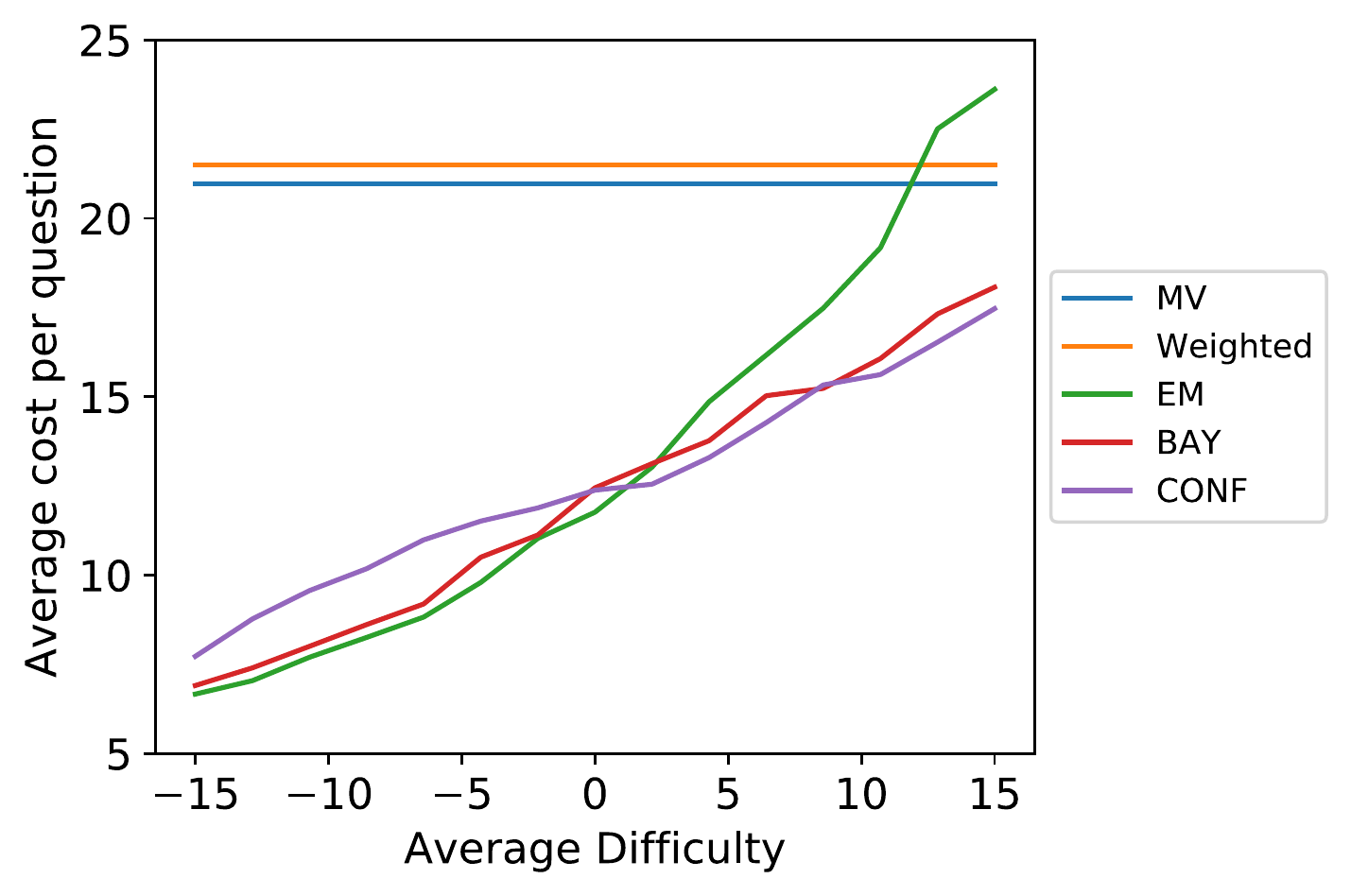}
    \label{fig:Difficultyb}
    \caption{Average Cost}
\end{subfigure}
\caption{Average Difficulty Test}
\label{fig:Difficulty}
\end{figure}

From Fig. \ref{fig:Difficulty} it can be seen that when the average difficulty for each question is increased, all the methods become less accurate. However, it can be noted that for the two Bayesian inference methods are more robust to the increase in difficulty as they do not lose accuracy as quickly. This could be because the harder questions force the systems to use more workers to maintain the confidence that is required. The ability for the probabilistic methods to realise that there is more disagreement in the system is what gives the probabilistic methods the edge over the voting methods. It allows the models to control how many workers need to be used to maintain the confidence, which allows for the cheaper labels when the questions are easier and trade-off the low cost for higher accuracy when there is more uncertainty in the label. Incorporating confidence here allows for more reliable labels on a wider range of difficulties. We can see that for the easier questions, the 5 methods perform similarly, but as the questions get harder, the difference between the probabilistic methods and the voting methods becomes clear.

\subsection{Adversarial Workers}
In this experiment, the effect of adding more adversarial workers was investigated. The number of adversaries was varied from 0 to 40, with 40 normal workers and 5 experts also in the worker pool.

\begin{figure}[!b]
\centering
\begin{subfigure}[b]{0.365\textwidth}
    \includegraphics[width=0.885\columnwidth]{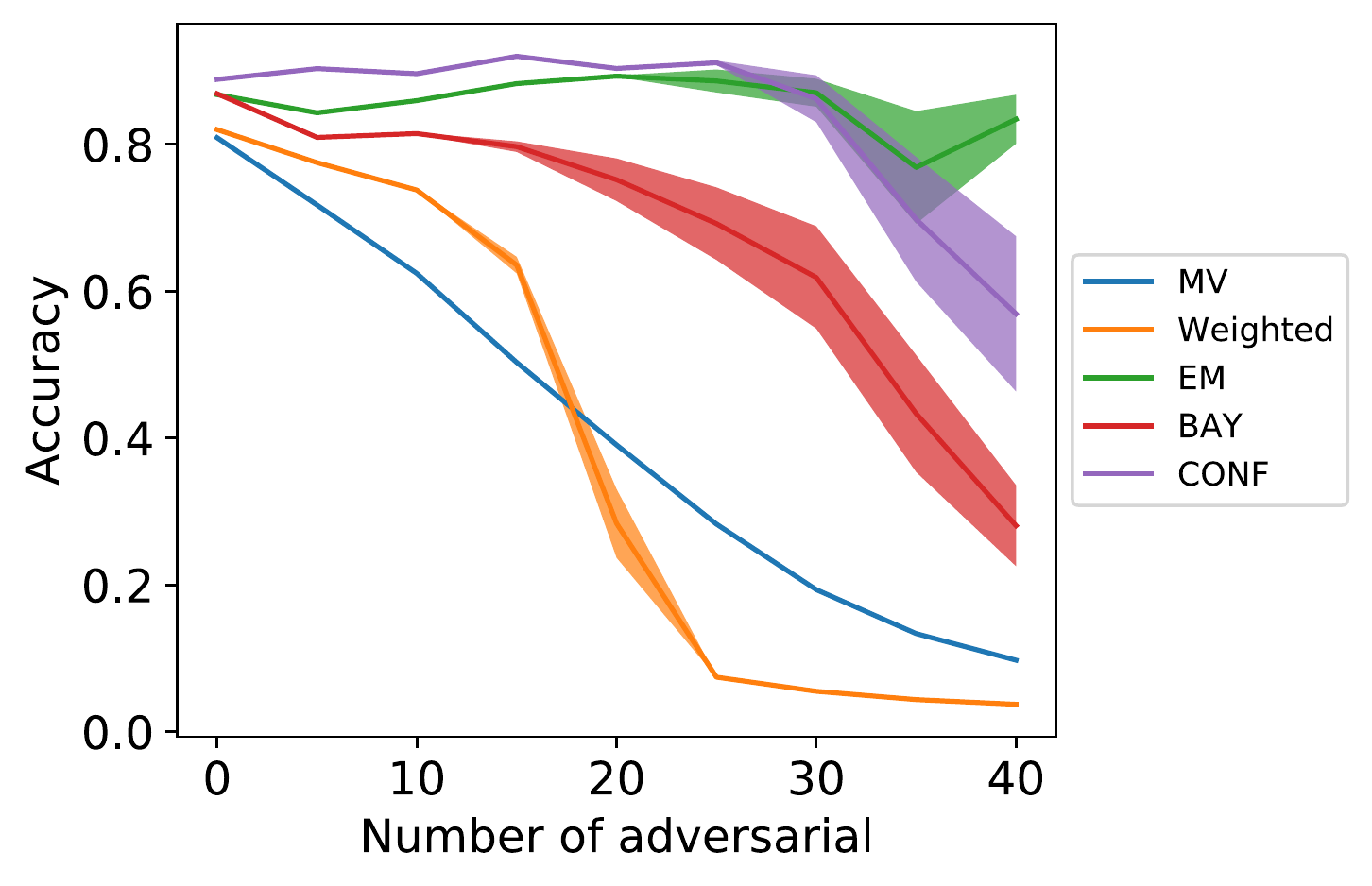}
    \label{fig:WorkerPoola}
    \caption{Accuracy}
\end{subfigure}

\begin{subfigure}[b]{0.365\textwidth}
     \includegraphics[width=0.885\columnwidth]{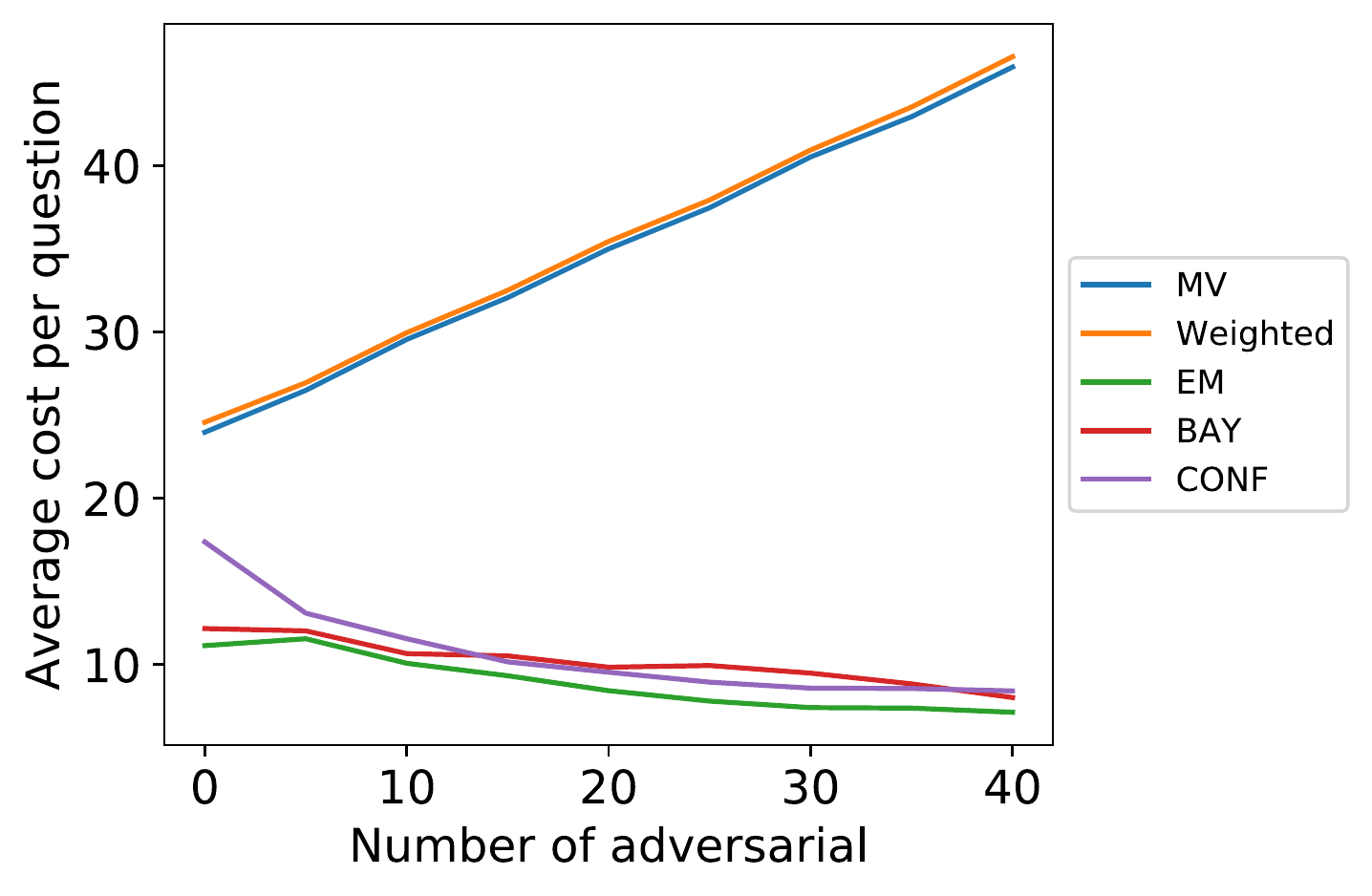}
    \label{fig:WorkerPoolb}
    \caption{Average Cost}
\end{subfigure}
\caption{Adversarial Test}
\label{fig:WorkerPool}
\end{figure}

In Fig. \ref{fig:WorkerPool} it can be seen how the different methods react towards having workers who are actively trying to sabotage the system. The maximum likelihood method and Bayesian inference with confidence update rule maintain their accuracy even when the worker pool is made up of around 35\% adversarial workers. The worst of the 5 methods was the weighted majority vote, which drops off rapidly as the number of adversaries increases, with its accuracy dropping below the majority vote model when the adversaries make up around 30\% of the system. The increase in cost for the voting methods can be attributed to the increase in worker pool size from adding more adversaries. It is worth noting that the variance in the results across the multiple test runs for the probabilistic methods increases greatly as the number of adversaries increase. This shows that the quality of the labels in these methods become more erratic, even though they generally still outperform the voting methods. 



\section{Conclusion and Future work}\label{sec:Conclusion}

Crowdsourcing is an invaluable tool in many data-driven fields. Because of the discrepancies between workers’ agreements, we often have to use multiple workers for each task. The majority vote performed the poorest of all the methods explored by only maintaining a competitive accuracy with the other methods for the easiest of questions. The weighted majority voting method improved performance slightly but still suffered from the high cost of the majority vote system. 

Of the three probabilistic methods, the Bayesian inference with the confidence update rule performed the best by having the highest average accuracy but at a slightly higher average cost. The different methods average costs tend to converge as more questions are asked making the Bayesian inference with the confidence update rule useful in situations where we consistently use the same set of workers to answer questions. In other cases when the workers are only being used for a small set of questions, the expectation-maximization method will be more beneficial with its lower average cost.

Future work should focus on testing the methods on real-world data, to relax some of the constraints imposed on the problem -- such as the independence of workers and questions -- and extend the system to non-binary classification problems.


\bibliographystyle{./bibliography/IEEEtran}
\bibliography{./bibliography/IEEEabrv,./bibliography/IEEEexample}

\end{document}